\documentclass[10pt,twocolumn,letterpaper]{article}

\usepackage[final]{iccv}      

\usepackage{url}            
\usepackage{booktabs}       
\usepackage{amsfonts}       
\usepackage{amsmath}
\usepackage{amsthm}
\usepackage{amssymb}
\usepackage{nicefrac}       
\usepackage{microtype}      
\usepackage{xcolor}         
\usepackage{lipsum}

\usepackage{graphicx}
\usepackage{multirow}

\usepackage{algorithm}
\usepackage{algpseudocode}

\usepackage{wrapfig}
\usepackage{xcolor} 
\usepackage{colortbl}
\makeatletter
\newcommand{\rmnum}[1]{\romannumeral #1}
\newcommand{\Rmnum}[1]{\expandafter\@slowromancap\romannumeral #1@}

\definecolor{iccvblue}{rgb}{0.21,0.49,0.74}
\usepackage[pagebackref,breaklinks,colorlinks,allcolors=iccvblue]{hyperref}

\def\paperID{15548} 
\def\confName{ICCV}
\def\confYear{2025}

\title{MMARD: Improving the Min-Max Optimization Process in Adversarial Robustness Distillation}

\author{Yuzheng Wang$^{1 }$ $\quad$
        Zhaoyu Chen$^{1}$ $\quad$
        Dingkang Yang$^{1 }$ $\quad$
        Yuanhang Wang$^{1}$ $\quad$
        Lizhe Qi$^{1,2 \dagger}$ \\ 
        \small$^1$Shanghai Engineering Research Center of AI \& Robotics, Academy for Engineering \& Technology, Fudan University \\
        \small$^2$Engineering Research Center of AI \& Robotics, Ministry of Education, Academy for Engineering \& Technology, Fudan University 
}

\begin{document}
\maketitle

\begin{abstract}
Adversarial Robustness Distillation (ARD) is a promising task to boost the robustness of small-capacity models with the guidance of the pre-trained robust teacher.
The ARD can be summarized as a min-max optimization process, \textit{i.e.}, synthesizing adversarial examples (inner) \& training the student (outer).
Although competitive robustness performance, existing ARD methods still have issues.
In the inner process, the synthetic training examples are far from the teacher's decision boundary leading to important robust information missing. 
In the outer process, the student model is decoupled from learning natural and robust scenarios, leading to the robustness saturation, \textit{i.e.}, student performance is highly susceptible to customized teacher selection.
To tackle these issues, this paper proposes a general Min-Max optimization Adversarial Robustness Distillation (MMARD) method. 
For the inner process, we introduce the teacher's robust predictions, which drive the training examples closer to the teacher's decision boundary to explore more robust knowledge.
For the outer process, we propose a structured information modeling method based on triangular relationships to measure the mutual information of the model in natural and robust scenarios and enhance the model's ability to understand multi-scenario mapping relationships.
Experiments show our MMARD achieves state-of-the-art performance on multiple benchmarks.
Besides, MMARD is plug-and-play and convenient to combine with existing methods.
\end{abstract}

\begin{figure*}[t]
	\centering
	\includegraphics[scale=0.385]{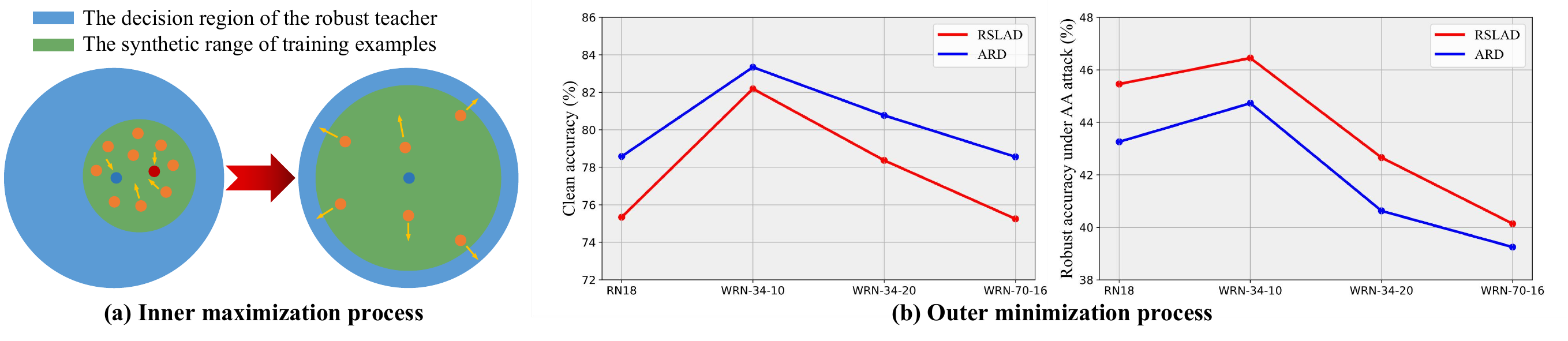}
	\caption{(a) A toy illustration about the synthetic range of existing methods (left) and expected range (right). The red dot denotes the real label, the blue dot denotes the teacher's prediction for the natural example, and the orange dots denote synthetic training examples.
    (b) A toy experiment about the students' clean and robust performance of ARD \cite{goldblum2020adversarially} and RSLAD \cite{zi2021revisiting} with various teachers.}
	\label{fig1}
\end{figure*}

\section{Introduction}
\label{sec1}

Deep learning technology has made significant progress in computer vision \cite{karras2019style,dosovitskiy2020image}, natural language processing \cite{devlin2018bert} and other fields \cite{radford2021learning}.
Deep neural networks (DNNs) are increasingly expected to be deployed in established and emerging fields of artificial intelligence.
However, many studies have shown that well-trained deep learning models are vulnerable to adversarial examples containing only minor changes \cite{szegedy2013intriguing,goodfellow2014explaining}.
This security risk raises concerns about model security in real-world applications, such as medical diagnosis \cite{ma2021understanding} and autonomous driving \cite{chen20203d}.

To address this challenge, many defensive strategies have been proposed \cite{madry2017towards,jia2019comdefend,kurakin2018adversarial,wang2020improving,wu2021wider,chen2022towards}.
Among them, Adversarial Training (AT) has been considered the most effective approach \cite{athalye2018obfuscated,croce2020reliable}.
AT is generally formulated as a min-max optimization process with inner maximization \& outer minimization.
In the inner maximization process, adversarial examples for model training (training examples) are synthesized from natural examples to discover robustness flaws in the model.
In the outer minimization process, model parameters are optimized with these examples to enhance adversarial robustness \cite{madry2017towards,wang2021convergence}.
Despite good robustness performance for large-capacity models, the help of AT for small models is very limited, \textit{i.e.}, the smaller the model, the poorer the robust performance \cite{gowal2020uncovering}, which hinders mobile deployment with limited memory and computational power.
This has motivated the combination of AT and knowledge distillation \cite{hinton2015distilling}.
The robustness of the small-capacity model is significantly improved with the help of pre-trained large-capacity robust models, a process called Adversarial Robustness Distillation (ARD) with a similar min-max optimization process.

Despite improving the robustness of small models, existing ARD methods still have issues in both inner maximization and outer minimization processes.
Specifically, for the inner maximization process, existing methods completely or partially ignore the robust information from the pre-trained teacher.
ARD \cite{goldblum2020adversarially}, IAD \cite{zhu2022reliable}, and MTARD \cite{zhao2022enhanced} only use the ground truth information and ignore the teacher's guidance.
RSLAD \cite{zi2021revisiting} introduces the teacher's guide for natural examples to improve performance.
However, for the robust model, the predictions for natural examples are farther away from the model's own decision boundary, which leads to the loss of vital information in the knowledge transfer process \cite{heo2019knowledge}.
For the outer minimization process, all existing methods heavily rely on the individual data example represented by the teacher in natural or robust scenarios, \textit{e.g.}, calculating Kullback-Leibler (KL) divergence loss between the student's and teacher's predictive distributions.
Just like calculating the cross-entropy loss between the student's predictions and one-hot real labels like AT process, these methods change the learning process from real labels (AT) to the teacher's soft labels, suffering from similar limitations of poor robustness for small models.
Therefore, existing methods contain a robust saturation issue, \textit{i.e.}, the student can only learn from a specific teacher framework \cite{zi2021revisiting}. 
Further increasing the capacity of the teacher model, the student's performance gradually deteriorates due to the larger capacity gap.

More intrigued, a toy illustration and a toy experiment are performed to validate the issues.
(\textbf{\rmnum{1}}) As shown in Figure~\ref{fig1} (a), the left figure shows synthetic adversarial examples in existing methods, \textit{i.e.}, they only use ground truth or teacher's predictions for natural examples, so a large number of examples near the teacher's decision boundary are lost, leading to the lack of robust information.
(\textbf{\rmnum{2}}) We choose two representative ARD methods \cite{goldblum2020adversarially,zi2021revisiting} to train the early 50 epochs on the CIFAR-10 \cite{krizhevsky2009learning} dataset with four robust teachers with various capacities \cite{croce2021robustbench} as shown in Figure~\ref{fig1} (b).
From left to right on the horizontal axis, the model capacities of the teachers increase gradually, \textit{i.e.}, ResNet (RN-18) \cite{he2016deep} and WideResNet (WRN-34-10, WRN-34-20 and WRN-70-16) \cite{zagoruyko2016wide}.
The students use RN-18 as the backbones. 
The clean accuracy and robust accuracy under AutoAttack (AA) attack \cite{croce2020reliable} show that the teachers' knowledge is not always efficient for the student due to capacity gap.
The observation confirms the robust saturation issue of existing ARD methods.

Based on these observations, we propose a novel method to improve the min-max optimization process in the ARD task called MMARD.
MMARD attempts to explore the training examples with more robust information, strengthen the student's adaptability to the teacher's model choice, and alleviate the robust saturation issue.
Firstly, the training examples close to the teacher's decision boundary help improve knowledge transfer efficiency \cite{heo2019knowledge}.
Based on this, in the inner maximization process, we introduce the teacher's robust predictions for the synthetic training examples.
The predictions will be continuously updated with the student's training process to constantly explore the differences between the teacher and student, which is different from the teacher's fixed predictions of natural examples or fixed real labels.
Secondly, in addition to individual data examples representation, the teacher's prediction representation between the natural and robust scenarios may also contain robust knowledge, \textit{e.g.}, how to avoid predicting pandas as bears.
Therefore, in the outer minimization process, we propose a triangular relationship among true labels, predictions in natural scenarios, and predictions in robust scenarios.
We think the relationship models the induction pattern of potential malicious attacks from the natural scenario to the robust scenario.
The pre-trained teacher model has learned certain invariant features in the samples to cope with sample perturbations. 
This invariant feature is universal and can help the student overcome the robustness saturation phenomenon, reducing the computational costs of customizing selecting models.
Besides, the MMARD is plug-and-play and convenient to combine with other methods and improve their performance.
Specifically, the primary contributions and experiments are summarized below:
\begin{itemize}
    \item We propose a novel method to improve the min-max optimization process in the ARD task.
    The inner maximization process of MMARD can be directly combined with other existing ARD methods to improve their performance.
    \item For the inner process, we add the training examples close to the teacher's decision boundary to explore the training examples with more robust information.
    For the outer process, we propose a triangular relationship to strengthen the student's adaptability to the teacher's model choice.
    \item Experimental results show that our MMARD improves model robustness and achieves state-of-the-art performance. 
    In particular, our method better overcomes the robust saturation issue in the ARD task.
\end{itemize}

\begin{table*}[t]
\centering
\caption{An overview of the learning objectives during min-max optimization in 7 defense methods. $\mathcal{L}_{min}$ is the loss function for the outer minimization while $\mathcal{L}_{max}$ is the loss function for the inner maximization. $S$ and $f$ are the optimized models. $T$ is the pre-trained teacher model. And $T_{nat}$ and $T_{adv}$ in MTARD are the teachers through normal training and AT, respectively. $\lambda$, $\alpha$, and $\beta$ are hyper-parameters in various methods. $\tau$ is the temperature constant.}
\setlength{\tabcolsep}{7mm}
\scalebox{1}{
\begin{tabular}{@{}c|c|c@{}}
\toprule
\textbf{Method} & $\mathcal{L}_{min}$ & $\mathcal{L}_{max}$ \\ \midrule
SAT    & $C\!E(f(x'),y)$ & $C\!E(f(x'),y)$ \\ \midrule
TRADES & $C\!E(f(x),y)+\lambda K\!L(f(x'),f(x))$ & $K\!L(f(x'),f(x))$ \\ \midrule
ARD    & $(1-\alpha )C\!E(S^\tau (x),y)+\alpha\tau ^{2} K\!L(S^\tau (x'),T^\tau (x))$ & $C\!E(S(x'),y)$ \\ \midrule
IAD    & $T_y(x')^\beta K\!L(S^\tau (x'),T^\tau (x))+(1-T_y(x')^\beta )K\!L(S^\tau (x'),S^\tau (x))$ & $C\!E(S(x'),y)$ \\ \midrule
RSLAD  & $(1-\alpha )K\!L(S(x),T(x))+\alpha K\!L(S(x'),T(x))$ & $K\!L(S(x'),T(x))$ \\ \midrule
MTARD  & $\alpha K\!L(S(x),T_{nat}(x))+(1-\alpha) K\!L(S(x'),T_{adv}(x'))$ & $C\!E(S(x'),y)$ \\ \midrule
\textbf{Ours}   & $K\!L(S(x'),T(x'))+\alpha l_\delta (\psi _t(T(x),y,T(x')), \psi _s(S(x),y,S(x')))$ & $K\!L(S(x'),T(x'))$ \\ \bottomrule
\end{tabular}
}
\label{tab1}
\end{table*}

\section{Preliminaries and Related Works}

\subsection{Adversarial Attack}

Adversarial attacks are malicious attacks on the data to perturb the predictions of the well-trained model \cite{szegedy2013intriguing}. 
The data tampered with by adversarial attacks is called adversarial examples, which refer to specially crafted input designed to look normal to humans but causes misclassification to a model \cite{kianpour2020timing,kumar2020adversarial}. 
Due to the primary cause of neural networks’ vulnerability to adversarial perturbation being their linear nature, CW \cite{carlini2017towards} uses an optimization-based adversarial examples synthesis method.
For faster synthesis, FGSM \cite{goodfellow2014explaining} uses a fast gradient sign method to synthesize adversarial examples quickly.
Based on this, PGD \cite{madry2017towards} adds an initial perturbation and uses multiple iterations to improve the performance.
The AutoAttack (AA) \cite{croce2020reliable} is a parameter-free, computationally affordable, and user-independent ensemble of four attacking methods.


\subsection{Adversarial Training and Adversarial Robustness Distillation}

Adversarial training is considered to be the most effective defense strategy against adversarial attacks \cite{athalye2018obfuscated,croce2020reliable}.
Recently, a large number of theories \cite{engstrom2019adversarial,zhang2019interpreting} and methods \cite{bai2021improving,guo2020meets} have proved the broad potential of AT.
The model can learn robust knowledge in advance by treating adversarial examples as data augmentation.
AT can be formulated as a min-max optimization process as follows:
$$
\underset{\mathrm {Outer \;\;   minimization}}{\underbrace{\mathrm {arg}\min\, \mathcal{L}_{min} (f(x'),y)}} ,
$$

\begin{equation}\label{eq1}
\mathrm {where}\;\; x'=\underset{\mathrm {Inner \;\;  maximization}}{\underbrace{  \underset{\left \| x'-x \right \|_p\le \epsilon  }{\mathrm{arg}\max}\,\mathcal{L}_{max}(f(x'),y) }}, 
\end{equation}
where $f$ is a training model, $y$ is the real labels, $x'$ is the training examples from natural examples $x$ within bounded $L_p$ distance $\epsilon$, $\mathcal{L}_{min}$ is the loss function for outer minimization process, and $\mathcal{L}_{max}$ is the loss function for inner maximization process.
Although Equation~\ref{eq1} allows the model to learn robust knowledge, the robustness performance is severely limited by the model capacity, \textit{i.e.}, the smaller the model capacity, the less significant robustness improvement.
For example, for lightweight ResNet-18 \cite{he2016deep} and MobileNet-V2 \cite{sandler2018mobilenetv2}, only relying on AT, the robust performance is struggling.
This issue hinders the deployment of robust DNN models to mobile devices with limited storage and computational resources, such as smartphones and driverless cars.

\begin{figure*}[t]
	\centering
	\includegraphics[scale=0.44]{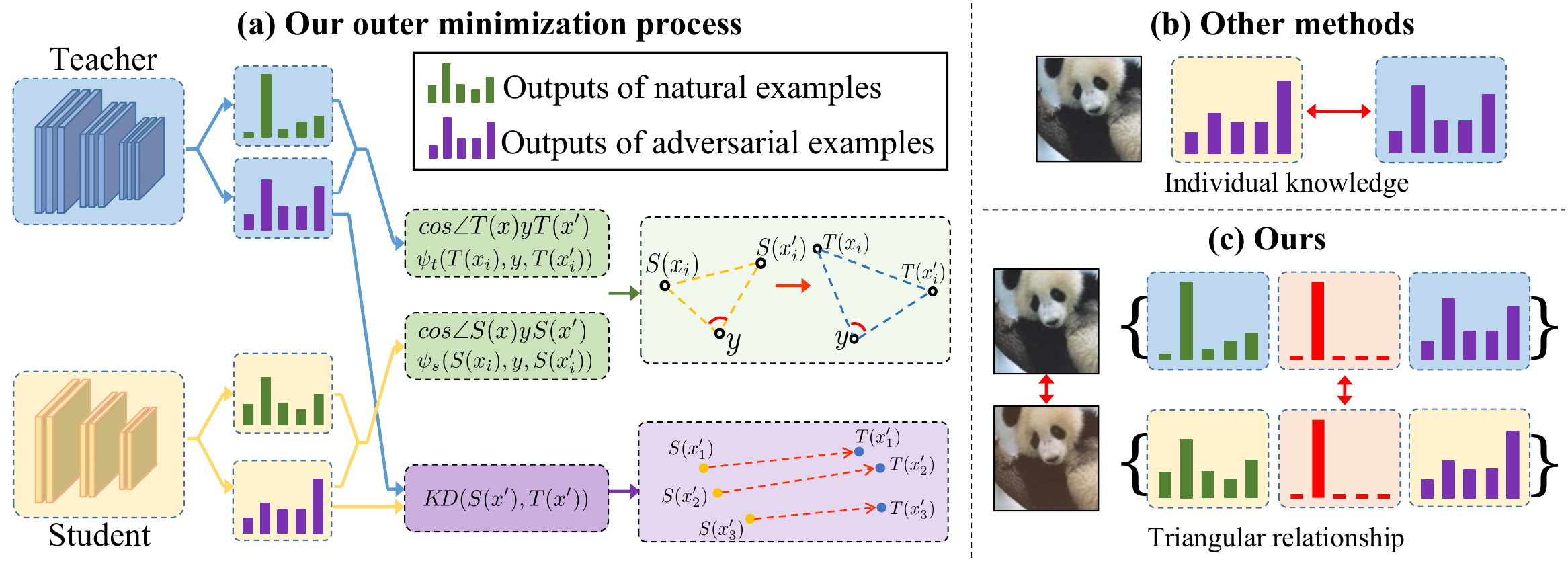}
	\caption{(a) The pipeline of our outer minimization process to train the student model. 
    In addition to the original individual example representation (purple box), we combine a triangular relationship to measure the models' mutual relations of the natural and robust scenarios (green box).
    (b) The existing individual data examples representation.
    The consistency regularization forces the student's prediction distribution close to the teacher's prediction distribution, which is susceptible to the influence of specific experimental settings.
    (c) Our triangular relationship of models' predictions on natural and corresponding adversarial examples.
    Our regularization transfers general invariant features to enhance the student's adaptability to different teachers.}
	\label{fig3}
\end{figure*}

Based on this, inspired by the insights of knowledge distillation \cite{hinton2015distilling}, the adversarial robustness of small models is significantly enhanced by distilling from large and robust models, a process known as Adversarial Robustness Distillation (ARD).
Similar to AT, the knowledge transfer process can also be formulated as a min-max optimization pipeline:
$$
x'= \underset{\left \| x'-x \right \|_p \le \epsilon  }{\mathrm{arg}\max}\, K\!L(S(x'),T(x)),
$$

\begin{equation}\label{eq2}
\underset{\mathrm {Outer \;\;   minimization}} {\underbrace{\mathrm {arg}\min\, \mathcal{L}_{min}}}\!\!\!,\; \mathrm {where}\;\; x'=\underset{\mathrm {Inner \;\;  maximization}}{\underbrace{  \underset{\left \| x'-x \right \|_p \le \epsilon  }{\mathrm{arg}\max}\,\mathcal{L}_{max}  }}. 
\end{equation}
As shown in Table~\ref{tab1}, we summarize the loss functions $\mathcal{L}_{min}$ and $\mathcal{L}_{max}$ used of representative AT methods (\textit{i.e.}, SAT \cite{madry2017towards} and TRADES \cite{zhang2019theoretically}) and state-of-the-art ARD methods (\textit{i.e.}, ARD \cite{goldblum2020adversarially}, IAD \cite{zhu2022reliable}, RSLAD \cite{zi2021revisiting}, and MTARD \cite{zhao2022enhanced}).
Specifically, SAT \cite{madry2017towards} adopts the original one-hot label and conducts supervised learning.
TRADES \cite{zhang2019theoretically} can be regarded as a self-distillation process, \textit{i.e.}, the representation of the same model in different scenarios as the teacher and student.
The adversarial robustness is significantly improved by consistent learning of natural and adversarial examples.
ARD \cite{goldblum2020adversarially} is the first adversarial robustness distillation method.
The robustness of small models is significantly improved due to insights from the pre-trained robust teacher model.
RSLAD \cite{zi2021revisiting} uses robust soft labels of the teacher model to make the robust knowledge transfer process more effective.
As the teacher model's accuracy on student-generated adversarial examples continues to decline, IAD \cite{zhu2022reliable} is a multi-stage strategy to allow student self-reflection in the later training process.
To balance clean and robust accuracy, MTARD \cite{zhao2022enhanced} uses a clean teacher and a robust teacher to train the student model simultaneously.

\section{Methodology}

For the current mainstream ARD methods, through the adversarial attack on the model to find complex training examples, the adversarial robustness of the model is continuously enhanced.
To tackle the issues mentioned in Sec.~\ref{sec1}, we optimize existing ARD methods from the perspective of inner maximization and outer minimization processes, respectively.
The complete algorithm is shown in Algorithm~\ref{alg1}.

\begin{figure}[h]
	\centering
	\includegraphics[scale=0.45]{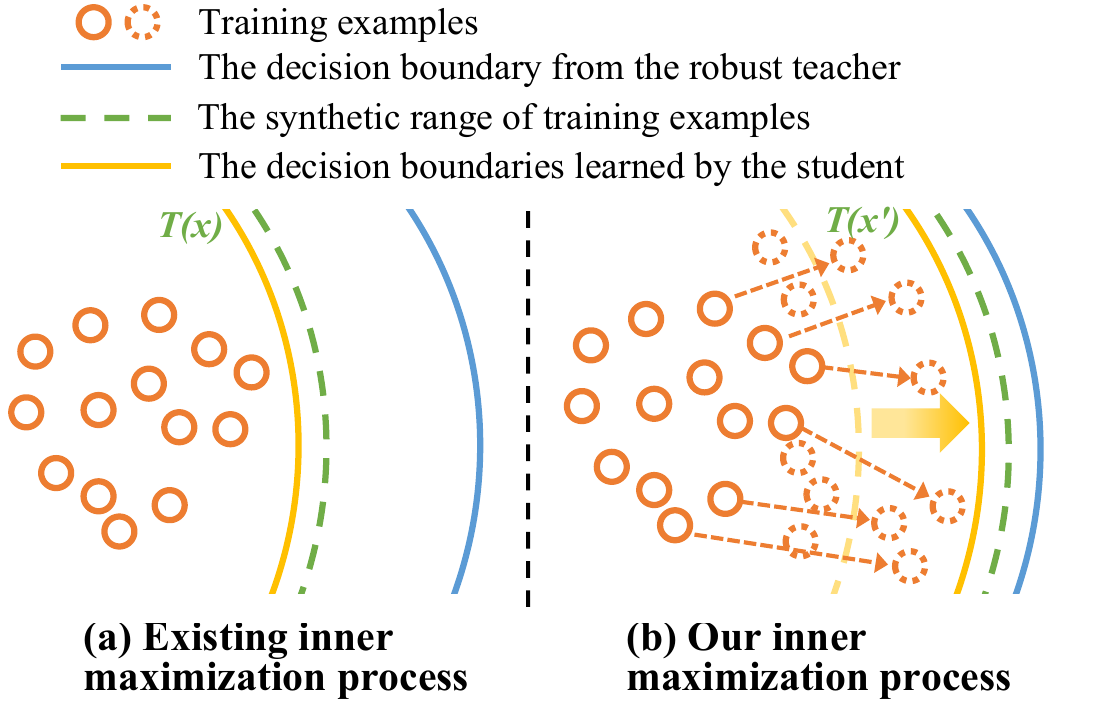}
	\caption{Diagrams of existing and our inner maximization process for synthesizing training examples. The training examples from existing methods are far from the teacher's decision boundary, containing less robust information.
    Our method reduces the discrepancy between the decision boundaries of the student and the teacher, thus promoting the robustness of the student.}
	\label{fig2}
\end{figure}

\subsection{Synthesizing Training Examples: Inner Maximization}

For the inner maximization process, we aim to synthesize the training examples with more robust information, determining the effectiveness of knowledge transfer.
Some studies have shown that the performance of a deep learning model largely depends on the difference between the decision boundary it learns and the real decision boundary between the actual class distribution \cite{lee1993feature,lee1997decision,he2018decision,karimi2019characterizing}.
Similarly, considering the requirements of adversarial robustness, the ARD task aims to make the student's prediction for adversarial examples not exceed its decision boundary.
Therefore, the decision boundary learned by the student directly determines its robust performance.
Besides, some studies on knowledge distillation theory have shown that the soft labels from the teacher can transfer information more effectively than the real one-hot label \cite{heo2019knowledge, zhang2022quantifying}.
According to the information entropy theory \cite{shannon1948mathematical}, the soft labels from the teacher contain more information about the decision boundary, thus enhancing student performance.
In this paper, we regard the ARD process as a student's robust decision boundary fitting optimization process with the insights from pre-trained robust teacher models.
Therefore, a critical point is \textit{how to synthesize training examples that contain more robust information to help the student learn better decision boundary}.

We attempt to optimize the inner maximization process in existing ARD methods by introducing the teacher's robust soft labels instead of fixed soft labels or real labels and exploring more information about the decision boundary to improve the student's performance.
For a robust teacher model trained by AT using adversarial examples, its predictions for natural examples are far from its own decision boundary.
Therefore, the existing methods lack part of the robust information (Figure~\ref{fig2} (a)).
The information contains more robust knowledge about the teacher's decision boundary, which is crucial and indispensable.
In contrast, our method further pulls in the student's and teacher's prediction distributions by synthesizing more informative training examples (Figure~\ref{fig2} (b)).
Our the inner maximization loss $\mathcal{L}_{max}$ can be calculated as:
\begin{equation}\label{eq3}
\mathcal{L}_{max}=K\!L(S(x'),T(x')),
\end{equation}
where $K\!L(,)$ denotes the Kullback-Leibler (KL) divergence loss. 
$S$ and $T$ denote the student and teacher, respectively.

\subsection{Training the Student: Outer Minimization}

For the outer minimization process, we aim to strengthen the student’s adaptability to the teacher’s model choice reducing the additional computational costs.
Existing ARD methods transfer robust knowledge through the individual data examples representation, \textit{e.g.}, constraining the agreement of student and teacher on the predicted distribution of training examples as:
\begin{equation}\label{eq4}
\mathcal{L}_{ARD}=K\!L(S(x'),T(x')).
\end{equation}

Essentially, Equation~\ref{eq4} is a kind of imitation learning for a single training example of natural or robust scenario, ignoring the implicit knowledge of the robust model for the natural and robust scenarios.
Although it allows the student to mimic the teacher's predictive distribution, the model capacity gap easily causes robust saturation phenomenon, \textit{i.e.}, the student's robustness does not increase continuously as the teacher's robustness increases but decreases due to the gap in the model capacity in some cases.
The phenomenon shows that the teacher's knowledge is not necessarily suitable for the student \cite{allen2020towards}.
On this basis, the student's performance is susceptible to the teacher's choice \cite{muller2019does,li2022asymmetric}.
This reliance on the custom teacher model adds additional computational costs for teacher's choice.
Additional invariance information \cite{schultz2003learning} is a promising approach to deal with this issue.
Several studies have shown that the teacher model has learned certain invariant features in the samples to cope with induced patterns of potentially malicious attacks ranging from the natural scenario to robust scenario.
These knowledge in knowledge distillation is general and invariant and can improve student performance \cite{yim2017gift,lee2018self}.
Therefore, another critical point is \textit{how to learn the knowledge of universal invariance in ARD task to strengthen the student’s adaptability to different teachers}.

\begin{algorithm}[t]
\caption{Training Process of ARD Method}
\label{alg1}
\begin{algorithmic}[1]
    \Require 
      A pre-trained robust teacher network $T$, a student $S$ with parameter $\theta$, training dataset $D=\left \{ (x_i,y_i) \right \}_{i=1}^{n}  $, learning rate for student $\eta$, learning rate for adversarial examples $\eta_{x'}$, batch size $m$, number of batches $M$, inner PGD steps $S$, inner maximum attack radius $\epsilon $, distillation epochs $N$, and the loss trade-off parameter $\alpha$.
    \State Initialize parameter $\theta$
    \For{epoch $=1$ \textbf{to} $N$}  
        \For{$\text{batch} =1$ \textbf{to} $M$}
            \State Sample a mini-batch $\mathcal{B} = \left \{ (x_i,y_i) \right \}_{i=1}^{m}$ from $D$
            
            \State \textit{\textbf{// Inner maximization}}
            
            \For{$x \in \mathcal{B}$}
            
                \State Construct adversarial examples as
                \State $x' \gets \prod_{\|x'-x\|_p \leq \epsilon}^{S} \Big[x' + \eta_{x'}\nabla_{\!x'}\mathcal{L}_{K\!L}\Big]$ 
                
                \State {$\mathcal{L}_{K\!L} = K\!L(S(x'), T(x'))$}
                
            \EndFor
            
            \State \textit{\textbf{// Outer minimization}}
            \State  Train the student $\theta \leftarrow \theta -\eta  \nabla_{\theta} \mathcal{L}_{min}   $
            \State $\mathcal{L}_{min} = \mathcal{L}_{ARD} + \alpha \mathcal{L}_{TRD}$

        \EndFor
    \EndFor
    \Ensure The robust student model  $S(\theta)$.
\end{algorithmic}
\end{algorithm}

\begin{table*}[t]
\centering
\caption{Teacher models used in our experiments and their performance.}
\setlength{\tabcolsep}{3mm}
\scalebox{1}{
\begin{tabular}{@{}ccccccccc@{}}
\toprule[1pt]
                Setting       & Dataset   & Teacher          & Clean   & FGSM    & PGD${\rm _{S}}$ & PGD${\rm _{T}}$ & CW      & AA      \\ \midrule
\multirow{2}{*}{Default} & CIFAR-10 & WideResNet-34-10 & 84.92\% & 60.87\% & 55.33\% & 56.61\% & 53.98\% & 53.08\% \\
                       & CIFAR-100 & WideResNet-70-16 & 60.86\% & 35.68\% & 33.56\%         & 33.99\%         & 31.05\% & 30.03\% \\ \midrule
\multirow{4}{*}{Other} & CIFAR-10  & ResNet-18        & 80.24\%   & 59.76\%   & 56.27\%           & 57.27\%           & 52.05\%   & 51.53\%   \\

& CIFAR-10 & WideResNet-34-10 & 84.92\% & 60.87\% & 55.33\% & 56.61\% & 53.98\% & 53.08\%  \\
                       & CIFAR-10  & WideResNet-34-20 &   88.70\%      &   63.26\%      &    55.91\%             & 57.74\%                &   54.76\%      &  53.26\%       \\
                       & CIFAR-10  & WideResNet-70-16 & 85.66\%   & 64.80\%   & 61.09\%           & 62.39\%           & 58.52\%   & 57.48\%   \\ \bottomrule[1pt]
\end{tabular}
}
\label{tab2}
\end{table*}

Based on this, we propose a triangular relationship to measure models' mutual relations of the natural and robust scenarios, as shown in Figure~\ref{fig3} (a).
Specifically, for a pre-trained teacher model $T$, we argue that its predicted distributions for both natural examples $T(x)$ and adversarial examples $T(x')$ contain relatively triangular knowledge to the real labels $y$.
We conduct an angle-wise measurement in the output representation space as:
$$
\psi_t =\cos\angle T(x) \,y\, T(x')=\left \langle e^{T(x)y},e^{T(x')y} \right \rangle,
$$
\begin{equation}\label{eq5}
\mathrm {where}\;\; e^{T(x)y}=\frac{T(x)-y}{\left \| T(x)-y \right \|_2 },e^{T(x')y}=\frac{T(x')-y}{\left \| T(x')-y \right \|_2 },  
\end{equation}
where $\psi_t$ denotes a triangular knowledge representation from the teacher.
Similarly, we can obtain the triangular knowledge representation of the student model as $\psi_s $.
Then, the triangular relationship distillation (TRD) loss is defined as:
\begin{equation}\label{eq6}
\mathcal{L}_{TRD} =\sum_{x \cup x' } l_\delta (\psi _t, \psi _s),
\end{equation}
where $l_\delta$ is the Huber loss. 
Unlike individual data examples representation, Equation~\ref{eq6} does not directly force the student to learn the teacher's prediction distribution but incorporates the robust information between the teacher's predictions on natural and corresponding adversarial examples. 
Combining the two-part robust knowledge as Equation~\ref{eq4} and \ref{eq6}, our overall outer minimization loss is computed as:
\begin{equation}\label{eq7}
\mathcal{L}_{min} = \mathcal{L}_{ARD} + \alpha \mathcal{L}_{TRD},
\end{equation}
where $\alpha$ is the loss trade-off parameter.

\section{Experiments}

\subsection{Experimental Setup}

\textbf{Dataset and Model.}
We conduct the experiments on two benchmark datasets in AT and ARD tasks, including 32$\times$32 CIFAR-10 and CIFAR-100  \cite{krizhevsky2009learning}.
For a fair comparison with other SOTA methods, we use the same pre-trained WideResNet (WRN) \cite{zagoruyko2016wide} teacher models (Table~\ref{tab2} Default) with RSLAD \cite{zi2021revisiting}.
In ablation experiments, we use various teacher models \cite{croce2021robustbench} (Table~\ref{tab2} Other) to test the student’s adaptability to the teacher’s model choice.
Besides, we select the ResNet-18 (RN-18) \cite{he2016deep} and MobileNet-V2 (MN-V2) \cite{sandler2018mobilenetv2} students following other ARD methods \cite{goldblum2020adversarially,zhu2022reliable,zi2021revisiting,zhao2022enhanced}.

\noindent \textbf{Baselines.}
We comprehensively compare our MMARD method with natural training method (Nature), representative adversarial training methods (SAT \cite{madry2017towards}, TRADES \cite{zhang2019theoretically}), and state-of-the-art adversarial robustness distillation methods (ARD \cite{goldblum2020adversarially}, IAD \cite{zhu2022reliable}, RSLAD \cite{zi2021revisiting}, and MTARD \cite{zhao2022enhanced}).
For ARD, the $\alpha$ is set to 1, following the vanilla version.
For MTARD, due to unpublished clean teachers, we report the original performance directly and ignore the method in other test experiments.
For other methods, we keep the default settings of the original papers.

\noindent \textbf{Implementation Details.}
All methods are implemented in PyTorch \cite{paszke2019pytorch} on RTX 3090 GPUs.
Following existing ARD methods, the inner maximization process uses a 10-step PGD (PGD-10) with a random start size of 0.001 and a step size of $2/255$ for the adversarial augmentation.
The perturbation bounds are set to $L_\infty $ norm $\epsilon=8/255$ for both datasets.
The outer minimization process uses Stochastic Gradient Descent (SGD) optimizer with a cosine annealing learning rate with an initial value of 0.1, momentum of 0.9, and weight decay of 2e-4 to train the student model.
We set the batch size to 128 and the total number of training epochs to 300.
Besides, our default trade-off parameter $\alpha$ is 1.
For natural training, we train the networks for 100 epochs on clean images.
The learning rate is divided by 10 at the 75-th and 90-th epochs.

\noindent \textbf{Attack Evaluation.}
We evaluate the adversarial robustness with five adversarial attacks: FGSM \cite{goodfellow2014explaining}, PGD${\rm _{S}}$ \cite{madry2017towards}, PGD${\rm _{T}}$ \cite{zhang2019theoretically}, CW$_\infty$ \cite{carlini2017towards} and AutoAttack (AA) \cite{croce2020reliable}.
These methods are the most commonly used for adversarial robustness evaluation.
The maximum perturbation is set as $\epsilon=8/255$.
The perturbation steps for PGD${\rm _{S}}$, PGD${\rm _{T}}$ and CW$_\infty$ are 20 epochs.
In addition, we test the accuracy of the models in normal conditions without adversarial attacks (Clean).

\subsection{Comparison with Other Methods}

\begin{table*}[t]
\centering
\caption{Adversarial robustness accuracy on CIFAR-10 with RN-18 and MN-V2 students. Bold numbers denote the best results.}
\setlength{\tabcolsep}{2mm}
\scalebox{0.865}{
\begin{tabular}{@{}c|cccccc|cccccc@{}}
\toprule[1pt]
\multirow{3}{*}{\textbf{Method}} & \multicolumn{6}{c|}{\textbf{RN-18}}           & \multicolumn{6}{c}{\textbf{MN-V2}}          \\ \cmidrule(l){2-13} 
                        & \multicolumn{6}{c|}{Attacks Evaluation} & \multicolumn{6}{c}{Attacks Evaluation} \\
                        & Clean  & FGSM  & PGD${\rm _{S}}$  & PGD${\rm _{T}}$  & CW  & AA & Clean  & FGSM  & PGD${\rm _{S}}$  & PGD${\rm _{T}}$  & CW & AA \\ \midrule
Nature                  &  \textbf{94.65\%}      &   19.26\%    &  0.0\%    &  0.0\%    &  0.0\%   & 0.0\%   &   \textbf{92.95\%}     &   14.47\%    &  0.0\%    &  0.0\%    & 0.0\%   &  0.0\%  \\
SAT                     &   83.38\%     & 56.41\%      &  49.11\%    &  51.11\%    & 48.67\%    & 45.83\%   &    82.48\%    &  56.44\%     &  50.10\%    &  51.74    & 49.33\%   & 46.32\%   \\
TRADES                  &  81.93\%      &  57.49\%     & 52.66\%     &  53.68\%    & 50.58\%    & 49.23\%   &   80.57\%     &  56.05\%     & 51.06\%     &  52.36\%    & 49.36\%   & 47.17\%   \\
ARD                     &   83.93\%     &  59.31\%     &  52.05\%    &  54.20\%    &  51.22\%   & 49.19\%   &  83.20\%      & 58.06\%      &  50.86\%    &  52.87\%    & 50.39\%  & 48.34\%   \\
IAD                     &  83.24\%      &  58.60\%     &  52.21\%    & 54.18\%     & 51.25\%    & 49.10\%   &  81.91\%      &  57.00\%     &  51.88\%    &  53.23\%    & 50.45\%   & 48.40\%   \\
RSLAD                   &    83.38\%    & 60.01\%      & 54.24\%     &  55.94\%    & 53.30\%    & 51.49\%   &   83.40\%     & 59.06\%      & 53.16\%     &  54.78\%    & 51.91\%   & 50.17\%   \\
MTARD                   &  87.36\%      &  \textbf{61.20\%}     &  50.73\%    &   53.60\%   &  48.57\%   &  $-$  &   89.26\%     & 57.84\%      &  44.16\%    &   47.99\%   & 43.42\%    & $-$   \\
\textbf{MMARD}                   &    84.63\%    &  60.19\%     &  \textbf{54.91\%}    & \textbf{56.51\%}     & \textbf{53.34\%}    & \textbf{52.50\%}    &   84.26\%     &   \textbf{59.28\%}    &  \textbf{53.70\%}    &  \textbf{55.32\%}    &  \textbf{52.28\%}   & \textbf{51.15\%}   \\ \bottomrule[1pt]
\end{tabular}
}
\label{tab3}
\end{table*}

\begin{table*}[t]
\centering
\caption{Adversarial robustness accuracy on CIFAR-100.}
\setlength{\tabcolsep}{2mm}
\scalebox{0.865}{
\begin{tabular}{@{}c|cccccc|cccccc@{}}
\toprule[1pt]
\multirow{3}{*}{\textbf{Method}} & \multicolumn{6}{c|}{\textbf{RN-18}}           & \multicolumn{6}{c}{\textbf{MN-V2}}          \\ \cmidrule(l){2-13} 
                        & \multicolumn{6}{c|}{Attacks Evaluation} & \multicolumn{6}{c}{Attacks Evaluation} \\
                        & Clean  & FGSM  & PGD${\rm _{S}}$  & PGD${\rm _{T}}$  & CW  & AA & Clean  & FGSM  & PGD${\rm _{S}}$  & PGD${\rm _{T}}$  & CW & AA \\ \midrule
Nature                  &  \textbf{75.55\%}      &   9.48\%    &  0.0\%    &  0.0\%    &  0.0\%   & 0.0\%   &   \textbf{74.58\%}     &   7.19\%    &  0.0\%    &  0.0\%    & 0.0\%   &  0.0\%  \\
SAT                     &   57.46\%     & 28.56\%      &  24.07\%    &  25.39\%    & 23.68\%    & 21.79\%   &    56.85\%    &  31.95\%     &  28.33\%    &  29.50    & 26.85\%   & 24.71\%   \\
TRADES                  &   55.23\%     &  30.48\%     & 27.79\%     & 28.53\%     & 25.06\%    & 23.94\%   &  56.20\%      &  31.37\%     & 29.21\%     &  29.83\%    & 25.06\%   &  24.16\%  \\
ARD                     &  60.64\%      & 33.41\%      & 29.16\%     & 30.30\%     & 27.85\%    & 25.65\%   &  59.83\%      & 33.05\%      & 29.13\%     & 30.26\%     & 27.86\%   & 25.53\%   \\
IAD                     &   57.66\%     & 33.26\%      & 29.59\%     & 30.58\%     & 27.37\%    & 25.12\%   & 56.14\%       & 32.81\%      &  29.81\%    & 30.73\%     & 27.99\%   & 25.74\%   \\
RSLAD                   &  57.74\%      & 34.20\%      & 31.08\%     & 31.90\%     & 28.34\%    & 26.70\%   &  58.97\%      & 34.03\%      & 30.40\%     & 31.36\%     & 28.22\%   & 26.12\%   \\
MTARD                   &   64.30\%     &   31.49\%    &  24.95\%    & 26.75\%     &  23.42\%   &  $-$  &  67.01\%      &  32.42\%     &  25.14\%    &  27.10\%    &  24.14\%  & $-$   \\
\textbf{MMARD}                   & 58.42\%    &  \textbf{34.56\%}     &  \textbf{31.36\%}     &    \textbf{32.05\%}     &    \textbf{28.94\%}  &  \textbf{27.21\%}  &  58.56\%       &    \textbf{34.45\%}      &     \textbf{30.95\%}  &    \textbf{31.96\%}    &   \textbf{28.27\%}   &   \textbf{26.53\%}  \\ \bottomrule[1pt]
\end{tabular}
}
\label{tab4}
\end{table*}

The robust performance of RN-18 and MN-V2 students is shown in Table~\ref{tab3} and \ref{tab4} on CIFAR-10 and CIFAR-100.
For the performance comparison, we select and report the best checkpoint of all methods among all epochs. 
The best checkpoints are based on the adversarial robustness performance against PGD${\rm _{T}}$ attack.

The experimental results demonstrate that our MMARD is competitive and achieves state-of-the-art robust accuracy under AA attack on various benchmarks.
For CIFAR-10, MMARD improves robust accuracy under AA attack 1.01\% and 0.98\% with RN-18 and MN-V2, respectively, compared to the previous SOTA method.
For CIFAR-100, MMARD improves by 0.51\% and 0.41\%, respectively.
While having higher robust performance, MMARD also has a higher clean accuracy than RSLAD in most cases.
It is worth noting that MMARD does not directly constrain the students' prediction distribution for natural examples. 
So the better clean accuracy also indicates that examples close to the teacher's decision boundary may also help the students in the natural scenario (further analysis can be as shown in Table~\ref{tab6}).

Meanwhile, we have some additional observations.
(\textbf{\rmnum{1}}) 
We find that the student performance of the ARD methods (ARD, IAD, RSLAD, and our MMARD) is significantly better than that of the AT methods (SAT and TRADES), regardless of clean accuracy or robust accuracy.
Such results demonstrate that the ARD methods can improve the performance of small models.
(\textbf{\rmnum{2}})
Although the student's capacity is smaller than that of the teacher in the ARD method, their performance is already very close.
For example, our trained RN-18 student model is only 0.29\% lower than the teacher on clean accuracy and 0.64\% lower on robust accuracy under CW attack on CIFAR-10.
This may be an essential factor in the limited improvement of MMARD.
(\textbf{\rmnum{3}})
Like MTARD, using clean and robust two teachers to transfer knowledge simultaneously can trade off the student's performance for the natural and robust scenarios.
With the help of a clean teacher, the clean accuracy of the students is significantly improved, which is significantly higher than other methods and the robust teacher.
However, the apparent drop in robust accuracy suggests the trade-off has no advantage for the robustness requirements in the ARD task.

\begin{table*}[t]
\centering
\caption{The performance (\%) of RN-18 students with teacher models of different capacities on CIFAR-10. \textbf{Average} represents the average performance of the same method with different teachers.}
\setlength{\tabcolsep}{1.6mm}
\scalebox{0.865}{
\begin{tabular}{@{}c|ccccc|ccccc@{}}
\toprule
\multirow{2}{*}{\textbf{Method}} & \multicolumn{5}{c|}{\textbf{Clean accuracy}}                & \multicolumn{5}{c}{\textbf{Robust accuracy under AA attack}} \\
                        & \textbf{RN-18} & \textbf{WRN-34-10} & \textbf{WRN-34-20} & \textbf{WRN-70-16} & \textbf{Average} & \textbf{RN-18}  & \textbf{WRN-34-10} & \textbf{WRN-34-20} & \textbf{WRN-70-16} & \textbf{Average} \\ \midrule
ARD   & 78.58 & 83.34 & 80.77 & 78.56 & 80.31 & 43.26 & 44.73 & 40.63 & 39.25 & 41.97 \\
RSLAD & 75.34 & 82.19 & 78.37 & 75.25 & 77.79 & 45.46 & 46.45 & 42.66 & 40.14 & 43.68 \\
\textbf{MMARD} & 77.54 & 83.35 & 86.27 & 85.09 & \textbf{83.06} & 47.02 & 47.76 & 47.13 & 47.43 & \textbf{47.34} \\ \bottomrule
\end{tabular}
}
\label{tab5}
\end{table*}

\subsection{Further Analysis}

\subsubsection{The Teacher’s Model Choice}

To compare the adaptability of students in various methods to different teachers, we select four teacher models from small to large capacity.
For a fair comparison, we choose ARD and RSLAD for 50 training epochs that do not require warming-up periods or an additional clean teacher model.
As shown in Table~\ref{tab5}, MMARD achieves significantly higher average performance in both natural and robust scenarios, improving by 5.27\% and 3.66\%, respectively.
For the other baseline methods, we find the robust saturation issue. 
Student performance drops when teachers choose better-performing WRN-34-20 and WRN-70-16.
Good performance of these methods requires customized teacher's model choice.
In contrast, our MMARD gains stable and reliable performance, strengthening the student's adaptability.

\begin{table}[t]
\centering
\caption{Combination of existing ARD methods with our inner maximization process as Equation~\ref{eq3}.}
\setlength{\tabcolsep}{3.4mm}
\scalebox{0.865}{
\begin{tabular}{@{}c|ccc@{}}
\toprule
\multirow{2}{*}{Method} & \multicolumn{3}{c}{Attacks Evaluation} \\
                        & Clean       & FGSM        & AA         \\ \midrule
ARD                     &   85.79 \textcolor{red}{(+1.86)}          &    60.05  \textcolor{red}{(+0.74)}         &  49.78 \textcolor{red}{(+0.59)}          \\
IAD                     &   85.45 \textcolor{red}{(+2.21)}         &    59.67 \textcolor{red}{(+1.07)}         &  49.46 \textcolor{red}{(+0.36)}          \\
RSLAD                   & 84.38 \textcolor{red}{(+1.00)}      & 60.06 \textcolor{red}{(+0.05)}      & 51.71 \textcolor{red}{(+0.22)}     \\ \bottomrule
\end{tabular}
}
\label{tab6}
\end{table}

\subsubsection{Other Methods with Our Inner Maximization}

To verify the influence of the optimization process in MMARD on other methods, we combine the inner maximization of MMARD with the outer minimization of other methods, \textit{i.e.},  $\mathcal{L}_{max}$ selects Equation~\ref{eq3} and $\mathcal{L}_{min}$ selects the original objective functions in Table~\ref{tab1}.
All RN-18 students are trained for 300 epochs on CIFAR-10.
Other settings are the same as Table~\ref{tab3}.
As shown in Table~\ref{tab6}, our inner maximization process can continue improving existing methods' performance regardless of natural or robust scenarios, compared to the vanilla versions. 
By synthesizing examples closer to the teacher's decision boundary, the information content of the training examples increases, enhancing student performance.
Notably, the clean accuracy of enhanced ARD and IAD (85.79\% \& 85.45\%) outperforms the robust teacher (84.92\%).
These two methods introduce the supervision information of real labels on the basis of improving the training examples information can significantly improve the learning of the natural scenario.

\begin{figure}[h]
	\centering
	\includegraphics[scale=0.65]{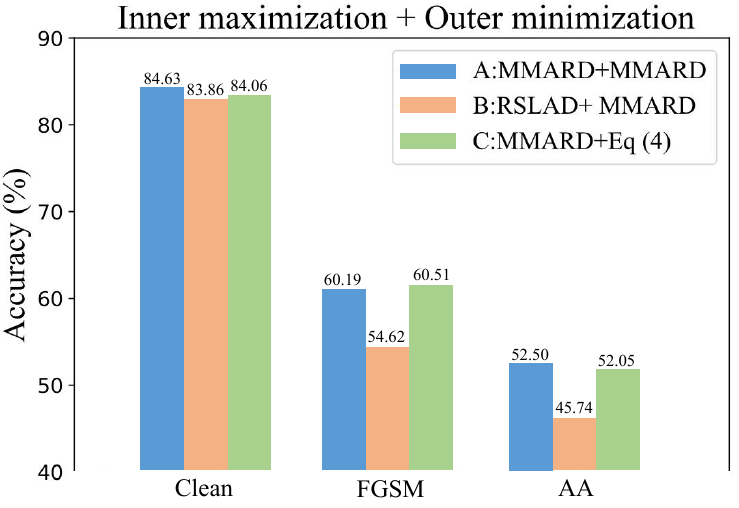}
	\caption{Students' performance about different combinations of inner maximization and outer minimization processes. RSLAD $\!\! + \!\!$ MMARD represents the combinations of RSLAD's inner process and MMARD's outer process. Eq (4) represents using only Equation~\ref{eq4} as the outer process.}
	\label{fig4}
\end{figure}

\subsubsection{Ablation Experiments on the Inner Maximization and Outer Minimization Process}

To further verify the effectiveness of our proposed min-max optimization process, we conduct ablation experiments about the combination of objective functions in the inner maximization and outer minimization processes.
Three min-max optimization strategies are conducted: (A) the vanilla MMARD, (B) the combination of RSLAD's inner process and MMARD's outer process, and (C) the combination of MMARD's inner process and Equation~\ref{eq4} as the outer process.
We train RN-18 students for 300 epochs on CIFAR-10.
The results are shown in Figure~\ref{fig4}.
An obvious conclusion is when the objective functions of the inner maximization are changed to the teacher's prediction of natural examples, the robust performance of the students drops significantly.
When a triangular relationship to measure models' mutual relations of the natural and robust scenario (Equation~\ref{eq6}) is ignored, the student's clean accuracy and robust accuracy under AA attack decrease slightly.
The results of these ablation experiments prove that our proposed min-max optimization process is effective.

\begin{table}[h]
\centering
\caption{Impact of the trade-off parameter $\alpha$.}
\setlength{\tabcolsep}{3mm}
\scalebox{0.865}{
\begin{tabular}{@{}c|cccccc@{}}
\toprule
\multirow{2}{*}{$\alpha$} & \multicolumn{6}{c}{Attacks Evaluation} \\
                   & Clean  & FGSM  & PGD${\rm _{S}}$  & PGD${\rm _{T}}$  & CW & AA \\ \midrule
0                  &  84.06      &   60.51    &  54.35    &  56.15    & 52.85   &  52.05  \\
0.5                &  84.57      &   \textbf{60.55}    & 54.85     & 56.24     & 53.17   & 52.17   \\
1                  &  \textbf{84.63}      &   60.19    & \textbf{54.91}     & \textbf{56.51}     & \textbf{53.34}   &  \textbf{52.50}  \\
2                  &  84.58      &  59.83     &  54.47    & 55.97     & 52.73   &  51.86  \\ \bottomrule
\end{tabular}
}
\label{tab7}
\end{table}

\subsubsection{Hyperparameter of the Trade-off Parameter $\alpha$}

We investigate the contributions of the loss trade-off parameter $\alpha$ in Equation~\ref{eq7} on student performance.
We train RN-18 students for 300 epochs on CIFAR-10.
The results are shown in Table~\ref{tab7}.
An appropriate value of $\alpha$ helps to improve students' clean and robust accuracy, \textit{e.g.}, $\alpha=0.5/1$, compared to entirely individual data examples representation ($\alpha=0$).
This observation also proves that our triangular relationship contributes to students' progress.

\section{Conclusion}

This paper proposes a plug-and-play MMARD method that improves the ARD task's inner maximization and outer minimization optimization process.
Firstly, the effectiveness of knowledge transfer is improved by synthesizing training examples closer to the teacher's decision boundary.
Then the student's performance is further improved by adding a triangular relationship between the teacher's predictions for the natural and robust scenarios.
Experimental results demonstrate that our MMARD outperforms other baselines and achieves new state-of-the-art performance.
Overall, MMARD better overcomes the robust saturation issue, alleviating reliance on custom teacher's model choice.


{
    \small
    \bibliographystyle{ieeenat_fullname}
    \bibliography{main}
}

\end{document}